\documentclass[runningheads]{llncs}
\usepackage{marvosym} 
\usepackage[T1]{fontenc}
\usepackage{graphicx}

\usepackage{float}
\usepackage{enumitem}
\usepackage{amsmath}
\usepackage[table]{xcolor}
\begin{document}
\raggedbottom
\title{Dynamic Orchestration of Multi-Agent System for Real-World Multi-Image Agricultural VQA}

\titlerunning{Dynamic Orchestration of Multi-Agent System}

\author{Yan Ke \and
Xin Yu\textsuperscript{(\Letter)}  \and
Heming Du\and
Scott Chapman\and
Helen Huang}
\authorrunning{Y. Ke et al.}
%
\institute{The University of Queensland, Brisbane QLD 4072, Australia \\
\email{\{y.ke,xin.yu,heming.du,scott.chapman,helen.huang\}@uq.edu.au}}
\maketitle              

\begin{abstract}
Agricultural visual question answering is essential for providing farmers and researchers with accurate and timely knowledge. However, many existing approaches are predominantly developed for evidence-constrained settings such as text-only queries or single-image cases. This design prevents them from coping with real-world agricultural scenarios that often require multi-image inputs with complementary views across spatial scales, and growth stages. Moreover, limited access to up-to-date external agricultural context makes these systems struggle to adapt when evidence is incomplete.  In addition, rigid pipelines often lack systematic quality control. To address this gap, we propose a self-reflective and self-improving multi-agent framework that integrates four roles, the Retriever, the Reflector, the Answerer, and the Improver. They collaborate to enable context enrichment, reflective reasoning, answer drafting, and iterative improvement.
 A Retriever formulates queries and gathers external information, while a Reflector assesses adequacy and triggers sequential reformulation and renewed retrieval. Two Answerers draft candidate responses in parallel to reduce bias. The Improver refines them through iterative checks while ensuring that information from multiple images is effectively aligned and utilized. Experiments on the AgMMU benchmark show that our framework achieves competitive performance on multi-image agricultural QA. 

\keywords{Large Multimodal Models  \and Multi-Agent Systems \and Retrieval Augmented Generation \and Agricultural Visual Question Answering}
\end{abstract}

\section{Introduction}
\label{sec:intr}
Large multimodal models (LMMs) have showcased impressive performance across diverse domains \cite{gpt3-2020}. Equipped with human-like language competence through training on massive data, these models have achieved a remarkable ability to understand and generate natural language. This capability has in turn spurred rapid adoption in fields such as healthcare and education. Despite their widespread utility, adapting language models to specific domains, particularly in agriculture, still requires further exploration. 

Although agricultural question answering has achieved rapid progress, current methods still encounter several critical challenges. To begin with, many existing systems \cite{rehman2023kisanqrs,koopman2024agask,yang2024shizishangpt} remain confined to text-only inputs or single-image cases and fail to exploit multimodal evidence such as crop images, growth records, multi-scale views, or temporal context, which are indispensable for practical agricultural scenarios. In addition, most approaches adopt a rigid pipeline without self-reflective or self-improving mechanisms \cite{silva2023gpt,liu2021efdet}. However, agricultural tasks often unfold under dynamic conditions such as variable growth stages, subtle disease symptoms, and shifting environments, where single-pass retrieval or reasoning can lead to unreliable outcomes, making iterative reflection essential. Moreover, current studies seldom incorporate systematic strategies for answer refinement \cite{yang2024application,xiong2025enhancing,zaremehrjerdi2025towards,chen2024adaptive}. This limits their capacity to handle ambiguous farmer queries, integrate field-specific observations, and generate recommendations that remain consistent with practical agricultural operations. In summary, prior work suffers from a narrow dependence on text-based or single-image inputs that are disconnected from real-world agricultural QA. It also lacks self-reflective reasoning for evolving crop and environmental conditions and offers inadequate strategies for refining answers into trustworthy guidance for farmers.

Building on these challenges, we design a self-reflective and  self-improving multi-agent framework that aims to provide more reliable and context-aware agricultural question answering. Instead of relying on narrow forms of input, the framework brings together diverse evidence. It combines temporal and spatial signals from several crop images, growth records and broader contextual resources such as weather information, agricultural policies, and scientific literature as needed. This integration allows the system to reason with knowledge that more closely mirrors the diverse information sources consulted in real agricultural practice. Because retrieval often returns noisy, partial, or stale evidence and agricultural queries are highly context dependent, we introduce a Reflector that continuously examines the adequacy of retrieved knowledge and intermediate outputs. Through iterative evaluation, the Reflector enables the system to adapt its reasoning to evolving crop and environmental contexts, reducing the risk of unreliable contexts. In addition, Answerer and Improver agents are responsible for drafting, cross-checking, and refining candidate responses. Their interaction establishes a systematic process of quality control. The Answerers generate diverse drafts in parallel, while the Improver iteratively refines them by aligning information from multiple images, integrating complementary details across views and stages, and ensuring that no single image dominates the reasoning process. This design produces outputs that are consistent, logically complete, and more faithful to the visual and contextual evidence needed for actionable recommendations in agricultural practice. Overall, the framework unifies context enrichment, self-reflection, answer drafting, and iterative improvement by coordinating these roles in an interactive workflow, thereby advancing both the robustness and practical value of agricultural QA. The main contributions of this paper are as follows:
\begin{itemize}[topsep=5pt, parsep=0pt, partopsep=0pt]
  \renewcommand\labelitemi{$\bullet$} 
  \item We propose a multi-agent framework for agricultural VQA, where Retriever, Reflector, Answerer, and Improver collaborate to move beyond rigid single-pass pipelines.  

  \item We introduce a self-reflective reasoning mechanism that evaluates the adequacy and relevance of evidence and supports query reformulation with renewed retrieval, which enables the system to adapt to dynamic agricultural contexts. 

  \item We design a self-improving process that iteratively checks and enhances responses, ensuring that information from multiple images is effectively aligned and utilized, which is crucial for real-world agricultural QA.  
\end{itemize}

\section{Related Work}
\subsection{Large Multimodal Models}
Pre-trained on large-scale text corpora, large language models (LLMs)~\cite{bert-2019,roberta-2019,gpt3-2020,zheng2023judging} have demonstrated remarkable performance across a broad spectrum of natural language processing tasks, such as text comprehension, generation, zero-shot transfer, and in-context learning. With billions of parameters, these models exhibit strong generalization capabilities. Nevertheless, traditional LLMs are constrained to the textual modality, which hinders their ability to process and reason over richer forms of input, including images, videos, and audio. To address this limitation, multimodal large language models have emerged, integrating LLMs with multimodal encoders or decoders to support inputs and outputs spanning multiple modalities. These models have been successfully applied to diverse tasks, including image-text understanding~\cite{liu2023llava,blip-2}, video-language modeling~\cite{lin2023video}. Recent studies also emphasize that current LMMs, despite their broad capabilities, still face fundamental challenges in complex multimodal reasoning \cite{shiri2024empiricalanalysisspatialreasoning}. In this work, we concentrate on agricultural image-text understanding, as it presents persistent challenges for LMMs.

\subsection{LLM-Empowered Agent} 
A variety of LLM-empowered agent frameworks have emerged to enhance reasoning and tool orchestration. LangChain \cite{langchain} provides a modular framework for connecting LLMs with external APIs and knowledge bases, becoming one of the most widely used toolkits. AutoGPT \cite{autogpt} demonstrates autonomous task execution by chaining reasoning steps and external tool calls. AutoGen \cite{wu2024autogen} supports multi-agent conversations, allowing customizable LLM collaborations for complex problem solving. CAMEL introduces role-playing strategies to coordinate cooperative behaviors between multiple agents. MetaGPT \cite{hong2024metagptmetaprogrammingmultiagent} formalizes task decomposition by embedding Standardized Operating Procedures, particularly for software engineering. While these frameworks show impressive orchestration capabilities, they are largely open-domain and require strong coding expertise, limiting their accessibility for domain-specific applications such as agriculture. 

\subsection{Agricultural Intelligent Question Answering}
Rehman et al. \cite{rehman2023kisanqrs} propose KisanQRS, a deep learning system achieving high accuracy on call log data, but it is limited to text-only queries. Silva et al. \cite{silva2023gpt} evaluate GPT-4 on agriculture exams and show strong performance, yet their study remains exam-focused without multimodal or agent-based extensions. Koopman et al. \cite{koopman2024agask} develop AgAsk for document retrieval with neural ranking, though it lacks multimodal reasoning and quality control.  Wei et al. \cite{wei2024snapdiagnoseadvancedmultimodal} demonstrate a retrieval paradigm that allows farmers to submit either textual symptom descriptions or plant images for disease identification. ShizishanGPT \cite{yang2024shizishangpt} integrates Retrieval-Augmented Generation, knowledge graphs, and external tools, but tested on a small crop-specific dataset. Zhang et al. \cite{zhang2024empowering} release the CROP benchmark to enhance crop science QA, while it remains restricted to rice and corn in text-only form. To address these gaps, we design a multi-agent agricultural QA framework that unifies iterative retrieval, dynamic agricultural information grounding, and quality-controlled answer synthesis, providing reliable outputs in knowledge-intensive agricultural contexts.

\section{Proposed Method}
\begin{figure}[!t]
\includegraphics[width=\textwidth]{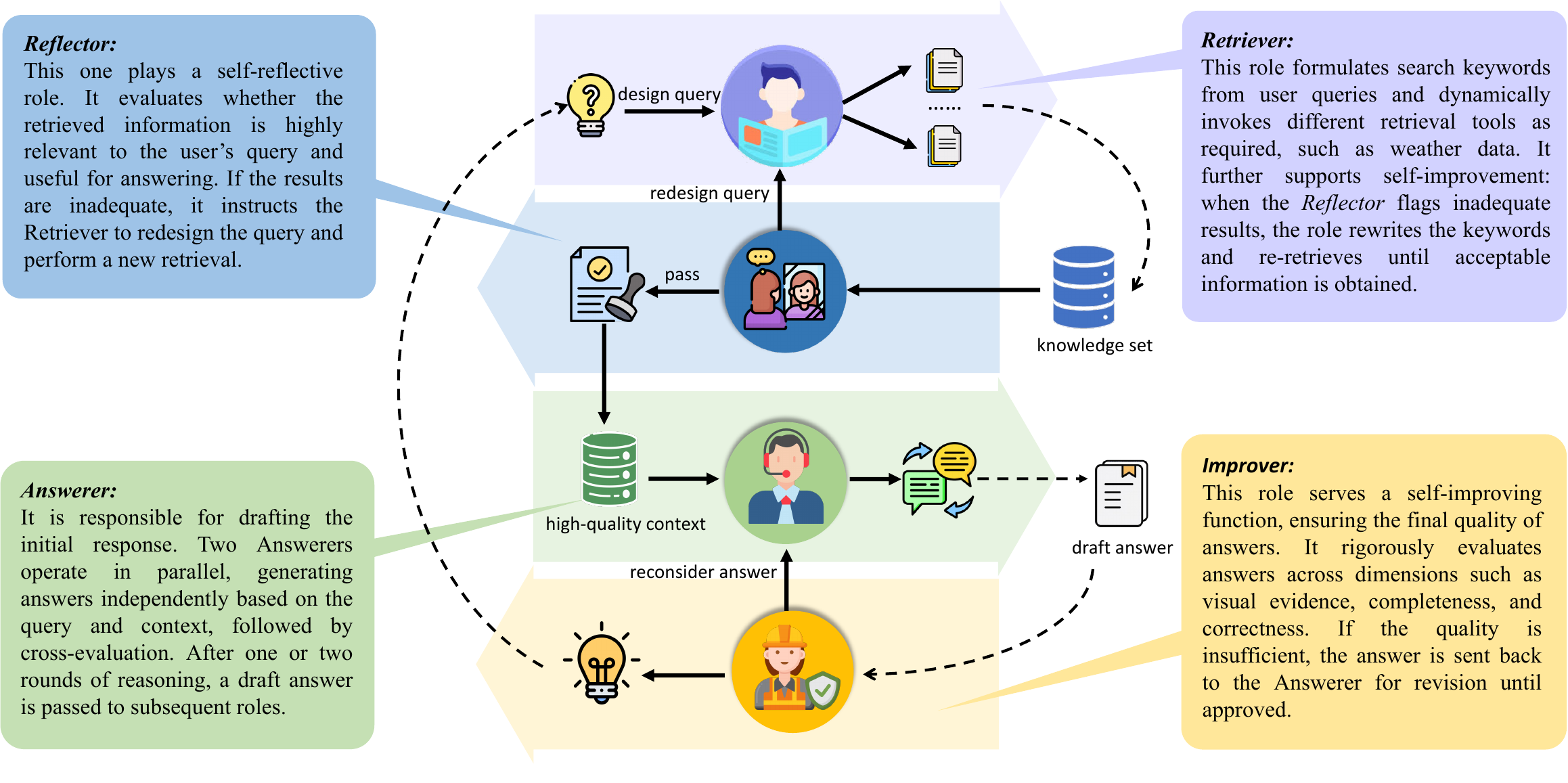} 
\vspace{-2.5em}
\caption{Multi-agent framework for agricultural question answering. The top-left illustrates the roles involved in the workflow. The middle row represents the practical workflow, including information retrieval, quality assurance, and expert deliberation. The bottom row depicts the control flow directed by the Manager.} \label{fig1}
\end{figure}  
Agricultural question answering requires handling context-sensitive knowledge, dynamic environmental factors, and effective utilization of visual inputs. To address these challenges, we propose a self-reflective and self-improving multi-agent framework that integrates context enrichment, reflective reasoning, answer drafting and iterative improvement. As shown in Fig.\ref{fig1}, the framework introduces four specialized agents with distinct responsibilities, including retrieving evidence, reflecting on its adequacy, drafting candidate answers, and improving final outputs. Through these mechanisms, the framework adapts to evolving agricultural conditions and maintains reliability across diverse scenarios. Section~3.1 presents the design of each agent role, and Section~3.2 describes how these roles interact within the overall workflow.

\subsection{Agent Role Design}
\setlist[itemize]{label=\textbullet} 

\begin{itemize}[leftmargin=*]
  \item\raisebox{-0.3em}{\includegraphics[height=1.5em]{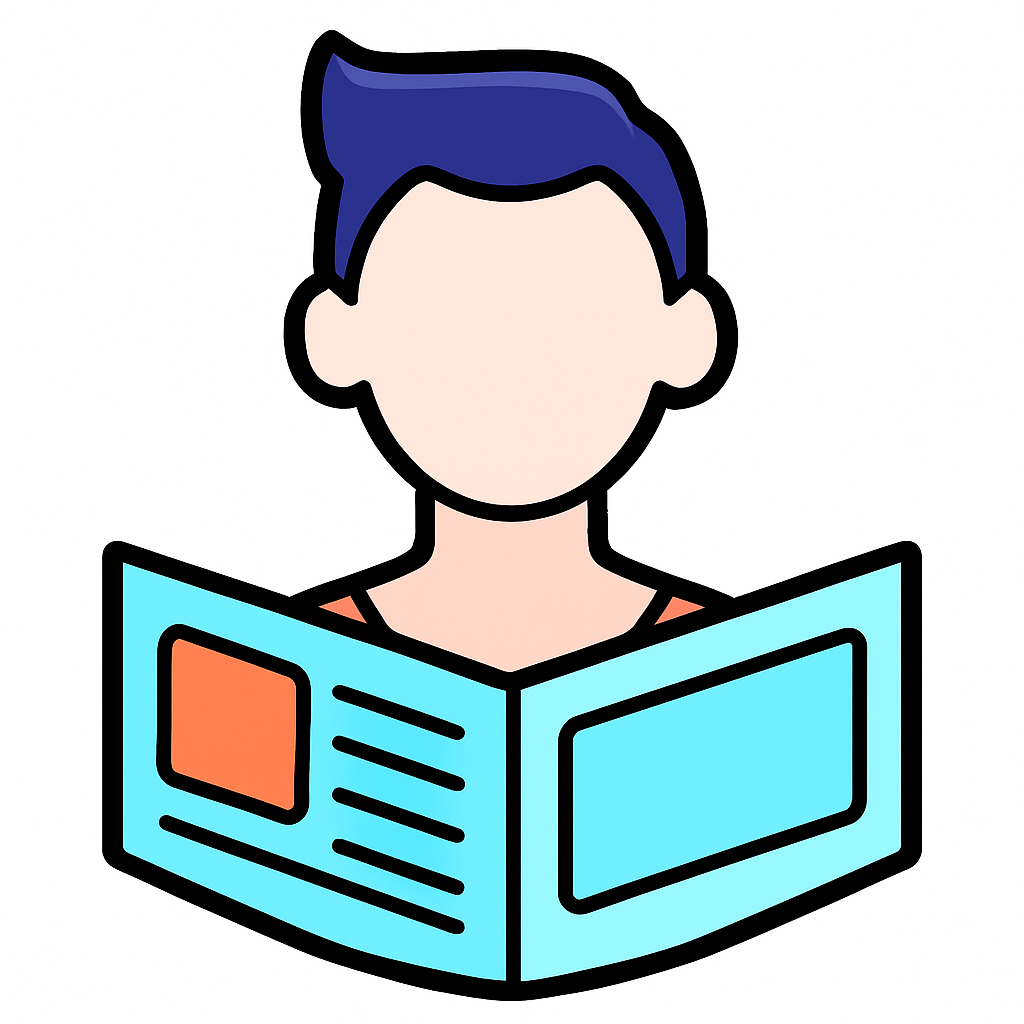}} \textbf{Retriever}. 
Retrieval-augmented generation substantially reduces hallucination by grounding language model outputs in external evidence \cite{sharma2024retrievalaugmentedgenerationdomainspecific}. Guided by this insight, we introduce the Retriever to ensure that agricultural QA is supported by robust and reliable information. Agricultural queries do not always require external resources, but in cases where context-sensitive or personalized problems demand precise and timely solutions, data such as weather updates, policy changes, and recent research become particularly valuable. The Retriever formulates search keywords from the user query and dynamically invokes different retrieval tools depending on the type of information required, such as local weather data or agricultural literature. It adapts to feedback from the Reflector by rewriting keywords and repeating the retrieval process until the evidence is sufficient. This iterative mechanism ensures that agricultural knowledge and situational data are collected. The retrieved content is then transferred to the Reflector for evaluation, providing a strong foundation for subsequent reasoning. In this way, the Retriever establishes a dynamic bridge between user queries and external evidence, enabling the system to reason on top of trustworthy agricultural context.
\end{itemize}

\begin{itemize}[leftmargin=*]
    \item\raisebox{-0.3em}{\includegraphics[height=1.5em]{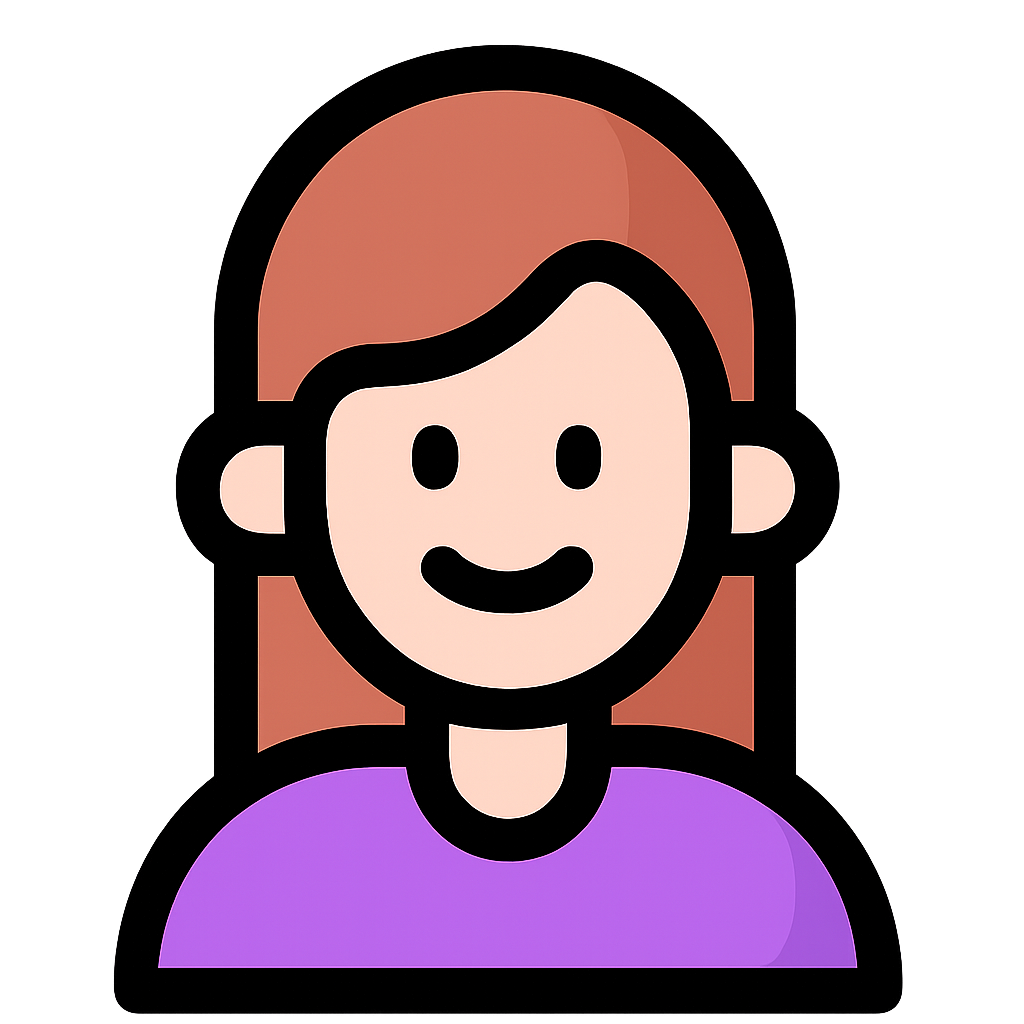}} \textbf{Reflector}. 
Retrieved documents often contain noise or irrelevant passages, which can weaken the reliability of downstream reasoning \cite{lewis2021retrievalaugmentedgenerationknowledgeintensivenlp}. Retrieval alone therefore does not guarantee that the information is fully relevant or adequate, which motivates the design of the Reflector. This role evaluates whether the retrieved evidence is appropriate for answering the query and prevents low-quality content from propagating forward. When deficiencies are identified, it instructs the Retriever to reformulate the query and conduct a new retrieval. The Reflector judges information quality from several aspects, such as topical relevance to the query, factual consistency across sources, and timeliness of the data. In agricultural QA, it further checks whether the evidence aligns with crop type, growth stage, and local environmental conditions, as generic or outdated knowledge can mislead decision-making. To operationalize this check, the Reflector applies weighted criteria and requires the evidence to surpass a predefined quality threshold before it is accepted. By enforcing this filtering process, the Reflector ensures that only reliable and context-aware information flows into the reasoning stage. Once the evidence reaches this threshold, the Reflector passes it to the Answerer to initiate drafting. This design closes the retrieval loop and guarantees that reasoning always begins with trustworthy agricultural context.
\end{itemize}

\begin{itemize}[leftmargin=*]
    \item\raisebox{-0.3em}{\includegraphics[height=1.5em]{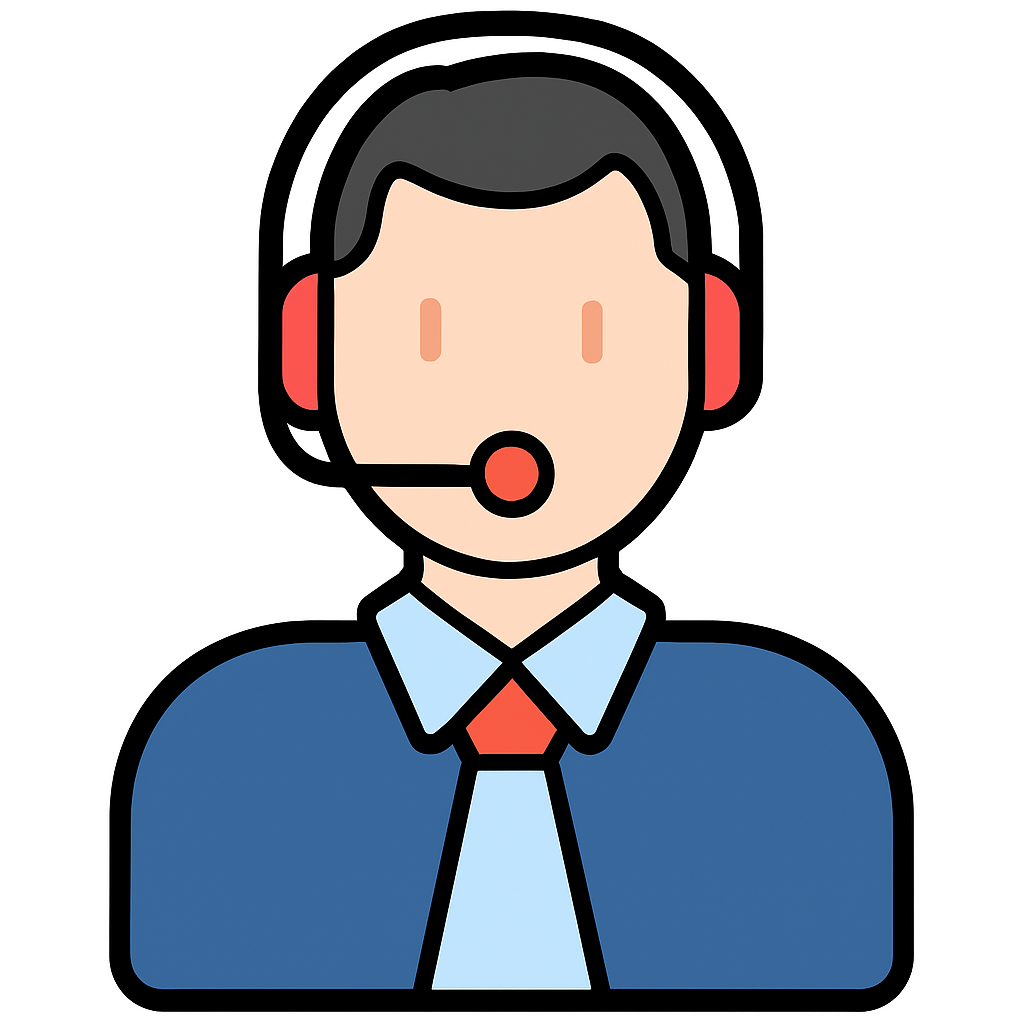}} \textbf{Answerer}. 
The Answerer follows the Reflector and is responsible for generating candidate answers from validated evidence. This role transforms the filtered context into draft responses that address the user query. Two Answerers operate in parallel and independently generate answers based on the same query and supporting information. They then cross-check and refine each other’s outputs, which helps reduce individual bias and capture complementary reasoning paths. In agricultural QA, this design is particularly useful because domain questions often admit multiple plausible interpretations, such as different symptom descriptions or alternative management practices. The collaborative process allows inconsistencies to be resolved and reasoning gaps to be filled. Through iterative refinement, a consolidated draft answer emerges that balances accuracy and comprehensiveness. This draft is then forwarded to the Improver for quality inspection, ensuring that the pipeline continues with a well-formed preliminary response.
\end{itemize}

\begin{itemize}[leftmargin=*]
    \item\raisebox{-0.3em}{\includegraphics[height=1.5em]{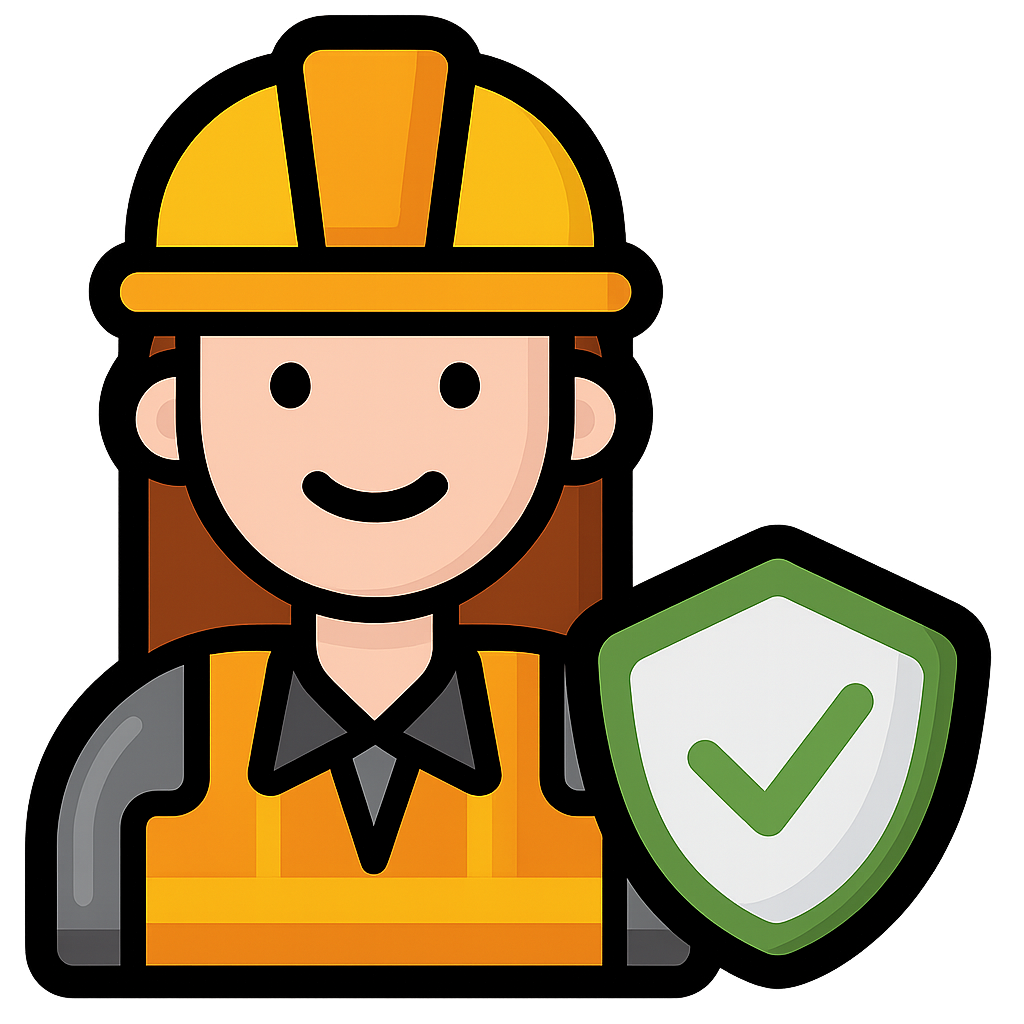}} \textbf{Improver}. 
Answer verification and quality control have been widely recognized as critical for reliable AI systems \cite{liu2021trustworthyaicomputationalperspective}. Ensuring the reliability of the final response therefore requires a role dedicated to quality assurance, which is fulfilled by the Improver. This role reviews the draft answer along dimensions such as completeness, instruction following, and consistency with visual evidence. For multi-image inputs, existing multimodal language models often rely on a single image while neglecting complementary information from the others \cite{wang2024muirbench}. Yet multi-image inputs are common in agricultural QA and are crucial for producing accurate answers. Based on these observations, the Improver evaluates whether responses are grounded in evidence from all images. When the evaluation reveals deficiencies, it guides the Answerers in further revision by aligning complementary details across views and growth stages, while also refining completeness and adherence to the user’s instruction as needed. The Improver applies structured evaluation, where each dimension contributes to an overall judgment of quality, and only when the accumulated assessment surpasses an acceptable threshold is the answer approved. By enforcing this procedure, the Improver strengthens the robustness and trustworthiness of the pipeline. 

\end{itemize}


\subsection{Collaboration Process}
Building on the role design introduced in the previous section, we now describe how these components collaborate to accomplish complex agricultural QA tasks. Previous work has often treated reasoning  and acting as separate processes, and this separation limits robustness and interpretability \cite{yao_react_2023}. Inspired by this work, our framework moves beyond a rigid pipeline by integrating reasoning and tool execution in an iterative loop, where the system reflects, invokes external resources, and adapts its actions based on intermediate outcomes. This design emulates flexible human problem-solving and alleviates prevalent issues such as hallucination and error propagation in chain-of-thought reasoning.

To formalize the collaboration process, we denote the query from the user at time $t$ as $q_t$ and the response from the answerer as $a_t$. The $s_t$ stands for the state at time $t$, including the dialogue history and intermediate outcomes. The decision process is represented by $P_{\text{decide}}$, which selects a query $q_t$ and a tool $T_t$ to invoke, conditioned on $s_t$. The execution of the retriever produces evidence $e_t$, whose relevance is judged by the reflector $G_{\text{reflect}}$. Candidate answers are then generated by the answerer according to $P_{\text{response}}$, and any disagreement between them is reconciled through the reconsideration function $f_{\text{reconsider}}$. Finally, the improver $G_{\text{answer}}$ validates the output. The decision process can be formulated as:

\begin{equation}
T_t \sim P_{\text{decide}}(T_t \mid s_t, q_t).
\end{equation}


The chosen tool $T_t$ executes on the query and returns evidence $e_t$:  

\begin{equation}
e_t = T_t(q_t).
\end{equation}

The reflector then evaluates the evidence. If it is judged insufficient, the query is reformulated and retrieval is repeated:  

\begin{equation}
g_t = G_{\text{reflect}}(e_t), \quad 
q_t \leftarrow 
\begin{cases}
q_t' & \text{if } g_t = \text{``no''}, \\
q_t  & \text{otherwise}.
\end{cases}
\end{equation}

Given the accumulated evidence $E_t = \{e_1, \dots, e_t\}$, two experts generate candidate answers independently:  

\begin{equation}
\{a_t^{(i)}\}_{i=1}^2 \sim P_{\text{response}}(a_t \mid s_t, E_t).
\end{equation}

If the two answers disagree, a reconsideration step is triggered to refine the responses:  

\begin{equation}
a_t' = f_{\text{reconsider}}\!\left(\{a_t^{(i)}\}_{i=1}^2, E_t\right).
\end{equation}

Finally, the improver evaluates the candidate answer. If it passes, the process terminates, otherwise it loops back to the answering step: 

\begin{equation}
y_t = G_{\text{improve}}(a_t'), \quad
\text{if } y_t = \text{``no''} \text{ then repeat } (4).
\end{equation}

A crucial feature of this process is its self-reflective and self-improving design, through which the system continually monitors and adjusts its own reasoning. The mechanism is implemented at both the retrieval and answer stages. At the retrieval stage, the reflector reviews retrieved documents and, when relevance is insufficient, triggers query rewriting and re-retrieval, effectively implementing a self-reflective mechanism. At the answer stage, responses are further checked to help mitigate hallucinations and ensure adherence to user intent. This validation is particularly important in agricultural VQA, where visual evidence often determines the correctness of an answer, so evaluation emphasizes not only completeness and accuracy but also grounding in image features. Through this two-layered quality control, the collaboration process enforces fidelity to evidence while promoting robustness and trustworthiness of the final answer.

\section{Experiments and Results}

\begin{table}[t]
\caption{Comparison of different models on agricultural VQA categories. Among them, DI, PI, SI, MI, and SD denote disease identification, pest identification, species identification, management instruction, and symptom description, respectively, while Average indicates the mean accuracy across all categories for each model.}\label{table:1}
\centering

\scriptsize                               
\renewcommand{\arraystretch}{1.25}        
\setlength{\extrarowheight}{1pt}          
\setlength{\tabcolsep}{5pt}               

\begin{tabular}{l|c c c c c c}
\hline
\multicolumn{1}{l|}{Category} & DI & PI & SI & MI & SD & Average \\ \hline
\rowcolor{gray!8}
\multicolumn{7}{l}{Proprietary Models} \\ \hline
GPT-4o           & 85.26 & 89.89 & 87.88 & 91.84 & 88.71 & 88.72 \\
Gemini-1.5-Pro   & 81.05 & 85.39 & 93.94 & 92.86 & 87.90 & 88.23 \\
Claude-3-Haiku   & 74.74 & 82.02 & 66.67 & 89.80 & 66.94 & 76.03 \\ \hline
\rowcolor{gray!8}
\multicolumn{7}{l}{Open-sourced Models} \\ \hline
Qwen2.5-VL-7B    & 74.74 & 78.65 & 82.83 & 85.71 & 84.68 & 81.32 \\
LLaVA-1.5-7B     & 66.32 & 62.92 & 67.68 & 74.49 & 77.42 & 69.77 \\
InternVL2-8B     & 45.26 & 39.33 & 46.46 & 81.63 & 70.16 & 56.57 \\ \hline
Ours    & 89.47 & 92.13 & 88.80 & 93.88 & 89.52 & 90.78 \\ \hline
\end{tabular}
\end{table}

\subsection{Experimental Setup}
\noindent\textbf{Dataset.}
To validate our method, we use AgMMU\cite{gauba2025agmmucomprehensiveagriculturalmultimodal}, built from 116k user–expert conversations across key categories such as disease, pest, and species identification, management instruction, and symptom description. The publicly available multiple-choice set includes 1,094 images with 1–10 per question, along with time and location context reflecting real-world scenarios.



\noindent\textbf{Models.}
For comparison, we evaluate representative proprietary and open-source multimodal models. The proprietary models include GPT-4o \cite{gpt4}, Gemini-1.5-Pro \cite{geminiteam2024gemini15unlockingmultimodal}, and Claude-3-Haiku \cite{anthropic2024claude3}, while the open-source models consist of Qwen2.5-VL-7B \cite{bai2025qwen25vltechnicalreport}, LLaVA-1.5-7B \cite{llava-v1.5}, and InternVL2-8B \cite{chen2025expandingperformanceboundariesopensource}.

\noindent\textbf{Evaluation Metrics.}
We evaluate predictions using the accuracy metric commonly adopted in QA, scoring 1 for correct answers and 0 otherwise, with overall accuracy as the average across all questions~\cite{gqa}.

\subsection{Evaluation across Agricultural VQA Tasks}
In this study, we conduct a comparative evaluation of agricultural question answering across five representative categories: disease identification, pest identification, species identification, management instruction, and symptom description. As shown in Table \ref{table:1}, the performance of different models varies considerably across the five categories of agricultural VQA. GPT-4o and Gemini-1.5-Pro achieve the strongest results among existing systems, both maintaining averages close to 89. Claude-3-Haiku performs less competitively, falling below 77, while Qwen2.5 reaches 81.32, surpassing Claude despite being an open-sourced model. LLaVA-1.5 and InternVL2, in contrast, remain substantially weaker, with averages under 70 and 60 respectively. 

Beyond model-specific scores, some general patterns can also be observed. All models tend to perform best on management instruction, where task requirements emphasize structured knowledge and reasoning, and all systems exceed 70 in this category. By contrast, disease identification proves more challenging, with consistently lower scores across models, reflecting the need for precise visual grounding and domain-specific knowledge. Our framework achieves the highest overall average of 90.78, with leading results in most categories and competitive performance. In particular, the improvements in disease and pest identification highlight the effectiveness of iterative retrieval and self-reflective answer validation in addressing categories that demand fine-grained agricultural expertise.


\subsection{Evaluation under Varying Image Settings}
\begin{table}[!t]
\caption{Comparison of different models on multi-image settings. ‘Num of Images’ denotes the number of images attached to each question, while ‘Distribution’ indicates the proportion of questions in each category.}
\label{table:2}
\centering

\scriptsize                               
\renewcommand{\arraystretch}{1.25}        
\setlength{\extrarowheight}{1pt}          
\setlength{\tabcolsep}{5pt}               

\begin{tabular}{lllll}
\hline
\multicolumn{1}{l|}{Num of Images}  & \multicolumn{1}{l|}{1-image} & \multicolumn{1}{l|}{2-image} & \multicolumn{1}{l|}{3-image} & $ \geq $4-image  \\ \hline
\multicolumn{1}{l|}{Distribution}   & \multicolumn{1}{l|}{38\%} & \multicolumn{1}{l|}{28\%} & \multicolumn{1}{l|}{30\%} & 4\%                  \\ \hline
\rowcolor{gray!8}
Proprietary Models                  &                              &                              &                              &                        \\ \hline
\multicolumn{1}{l|}{GPT-4o}         & \multicolumn{1}{l|}{88.02}   & \multicolumn{1}{l|}{90.78}   & \multicolumn{1}{l|}{86.84}   & 95.00                  \\
\multicolumn{1}{l|}{Gemini-1.5-Pro} & \multicolumn{1}{l|}{88.02}   & \multicolumn{1}{l|}{87.23}   & \multicolumn{1}{l|}{87.50}   & 100.00               \\
\multicolumn{1}{l|}{Claude-3-Haiku} & \multicolumn{1}{l|}{71.88}   & \multicolumn{1}{l|}{78.72}   & \multicolumn{1}{l|}{75.00}   & 90.00                  \\ \hline
\multicolumn{1}{l|}{Average}        & \multicolumn{1}{l|}{82.64}   & \multicolumn{1}{l|}{85.58}   & \multicolumn{1}{l|}{83.11}   & 95.00                  \\ \hline
\rowcolor{gray!8}
\multicolumn{5}{l}{Open-sourced Models} \\ \hline
\multicolumn{1}{l|}{Qwen2.5-VL-7B}  & \multicolumn{1}{l|}{82.29}   & \multicolumn{1}{l|}{82.27}   & \multicolumn{1}{l|}{79.61}   & 85.00                  \\
\multicolumn{1}{l|}{LLaVA-1.5-7B}   & \multicolumn{1}{l|}{73.44}   & \multicolumn{1}{l|}{68.79}   & \multicolumn{1}{l|}{66.45}   & 80.00                  \\
\multicolumn{1}{l|}{InternVL2-8B}   & \multicolumn{1}{l|}{51.56}   & \multicolumn{1}{l|}{60.99}   & \multicolumn{1}{l|}{60.26}   & 75.00                  \\ \hline
\multicolumn{1}{l|}{Average}        & \multicolumn{1}{l|}{69.10}   & \multicolumn{1}{l|}{70.68}   & \multicolumn{1}{l|}{68.77}   & 80.00                  \\ \hline
\multicolumn{1}{l|}{Ours}  & \multicolumn{1}{l|}{91.67} & \multicolumn{1}{l|}{90.07} & \multicolumn{1}{l|}{90.79} & 95.00 \\ \hline
\end{tabular}%
\end{table}

Agricultural visual question answering often depends on careful inspection of crops, pests, and disease symptoms, where multiple images taken from different perspectives or growth stages provide complementary evidence. While multi-view and multi-stage resources can substantially improve recognition accuracy, they also pose challenges for model generalization across diverse conditions \cite{shen2024mmwlauslanmultiviewmultimodalwordlevel}. To better understand these dynamics, it is necessary to examine model performance under varying numbers of input images, which offers deeper insight into the capacity of different approaches to exploit multi-view information effectively. 

\begin{figure} [!htbp] 
\includegraphics[width=\textwidth]{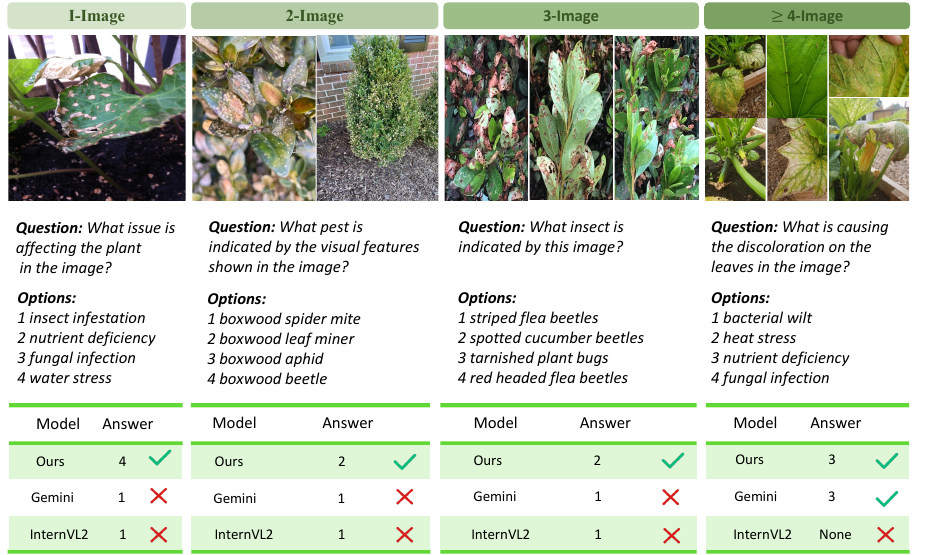}
\caption{Examples of agricultural visual question answering across different numbers of images. Each example shows the input images, corresponding question with multiple-choice options, and model predictions.} \label{fig2}
\end{figure}  

As shown in Table \ref{table:2}, our model maintains high robustness across all conditions, consistently exceeding 90 regardless of the number of images. On single-image and three-image questions, which together make up more than two-thirds of the dataset, our framework achieves the best scores. For two-image cases, GPT-4o slightly outperforms with 90.78. In the at least 4-image category, Gemini attains 100, whereas our method reaches 95, remaining close to the top and well above all open-sourced baselines.

Across models, a general pattern is that all systems reach their best performance in the $\geq$ 4 setting. This suggests that multiple images provide richer cues by revealing diverse visual features, such as different angles of a leaf or sequential stages of disease progression. Proprietary models generally show stronger capacity for leveraging multi-image information, though Qwen2.5 demonstrates competitive results against weaker proprietary systems. Our framework achieves strong performance overall in the multi-image setting underscoring its ability to integrate cross-image evidence effectively for agricultural VQA. 

We further compare model performance on agricultural VQA tasks with varying numbers of input images as shown in Fig.
\ref{fig2}. To this end, we select representative VQA triples from real-world user scenarios involving multi-view observations. The question types primarily cover common tasks such as pest identification and disease diagnosis. The results highlight the robustness of our framework in addressing complex real-world visual contexts for agricultural decision support.

\section{Conclusion}
In this work, we address the challenges of agricultural visual question answering, where real applications demand multimodal evidence, multi-image reasoning and systematic quality control. We present a self-reflective and self-improving multi-agent framework with four specialized roles that collaborate through context enrichment, self-reflection, answer drafting, and iterative improvement. This design enables the system to supplement incomplete knowledge and to make full use of multi-image inputs, avoiding reliance on a single view and providing answers that are reliable and practical. Experiments on the AgMMU benchmark confirm that the framework delivers competitive performance in multi-image agricultural QA while mitigating issues such as single-image fixation and weak cross-image integration. We believe this work represents a promising direction for enabling more reliable and context-aware question answering systems in agriculture, and it opens avenues for extending multi-agent strategies to agricultural question answering systems in future work.



%
%
%
%
\bibliographystyle{splncs04}
\bibliography{reference}




\end{document}